\documentclass[10pt,twocolumn,letterpaper]{article}

\usepackage{cvpr}
\usepackage{times}
\usepackage{epsfig}
\usepackage{graphicx}
\usepackage{amsmath}
\usepackage{amssymb}
\usepackage{color}

\usepackage{amsfonts} 
\usepackage{amsmath}
\usepackage{capt-of,etoolbox}
\usepackage{multirow}
\usepackage{booktabs}
\usepackage{diagbox}

\usepackage[pagebackref=true,breaklinks=true,letterpaper=true,colorlinks,bookmarks=false]{hyperref}

\cvprfinalcopy 

\def\cvprPaperID{6680} 

\makeatletter
\apptocmd\@maketitle{{\myfigure{}\par}}{}{}
\makeatother

\begin{document}
\vspace{-5mm}
\newcommand\myfigure{
\centering
    \includegraphics[width=0.95\linewidth]{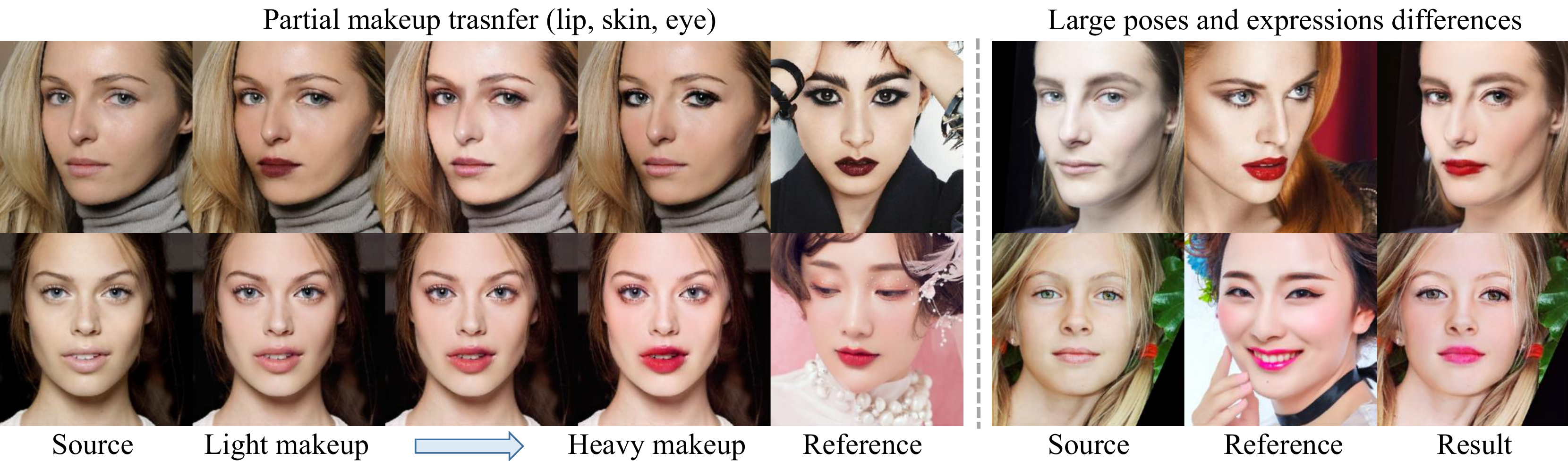}
\captionof{figure}{Our model allows users to control both the shade of makeup and facial parts to transfer. The first row on the left shows the results of only transferring partial makeup style from the reference. The second row shows the results with controllable shades. Moreover, our method can perform makeup transfer between images that have different poses and expressions, as shown on the right part of the figure. Best viewed in color.}
\label{p1}
}

\title{PSGAN: Pose and Expression Robust Spatial-Aware GAN for Customizable Makeup Transfer}


\author{Wentao Jiang\textsuperscript{\rm 1}, Si Liu\textsuperscript{\rm 1}, Chen Gao\textsuperscript{\rm 2,4}, Jie Cao\textsuperscript{\rm 3,4}, Ran He\textsuperscript{\rm 3,4}, Jiashi Feng\textsuperscript{\rm 5}, Shuicheng Yan\textsuperscript{\rm 6} \\ 
\textsuperscript{\rm 1}School of Computer Science and Engineering, Beihang University \\
\textsuperscript{\rm 2}Institute of Information Engineering, Chinese Academy of Sciences \\
\textsuperscript{\rm 3}Institute of Automation, Chinese Academy of Sciences \\
\textsuperscript{\rm 4}University of Chinese Academy of Sciences 
\textsuperscript{\rm 5}National University of Singapore 
\textsuperscript{\rm 6}YITU Tech \\
{\tt\small \{jiangwentao, liusi\}@buaa.edu.cn, gaochen@iie.ac.cn, jie.cao@cripac.ia.ac.cn} \\  {\tt\small rhe@nlpr.ia.ac.cn, elefjia@nus.edu.sg, shuicheng.yan@yitu-inc.com}
}

\maketitle
 
\begin{abstract} 
    In this paper, we address the makeup transfer task, which aims to transfer the makeup from a reference image to a source image. Existing methods have achieved promising progress in constrained scenarios, but transferring between images with large pose and expression differences is still challenging. Besides, they cannot realize customizable transfer that allows a controllable shade of makeup or specifies the part to transfer, which limits their applications. To address these issues, we propose Pose and expression robust Spatial-aware GAN (PSGAN). It first utilizes Makeup Distill Network to disentangle the makeup of the reference image as two spatial-aware makeup matrices. Then, Attentive Makeup Morphing module is introduced to specify how the makeup of a pixel in the source image is morphed from the reference image. With the makeup matrices and the source image, Makeup Apply Network is used to perform makeup transfer. Our PSGAN not only achieves state-of-the-art results even when large pose and expression differences exist but also is able to perform partial and shade-controllable makeup transfer. We also collected a dataset containing facial images with various poses and expressions for evaluations.
\end{abstract}
 
\vspace{-4mm}
\section{Introduction}

We explore the makeup transfer task, which aims to transfer the makeup from an arbitrary reference image to a source image. It is widely demanded in many popular portrait beautifying applications.
Most existing makeup transfer methods \cite{Li2018BeautyGANIF,ChenBeautyGlow2019,Chang2018PairedCycleGANAS,Gu2019LADNLA} are  based on Generative Adversarial Networks (GANs) \cite{Goodfellow2014GenerativeAN}. They generally use face parsing maps and/or facial landmarks as a preprocessing step to facilitate the subsequent processing and adopt the framework of CycleGAN \cite{Zhu2017UnpairedIT} which is trained on unpaired sets of images, i.e., non-makeup images and with-makeup images.

 
However, existing methods mainly have two limitations. 
Firstly, they only work well on frontal facial images with neutral expression since they lack a specially designed module to handle the misalignment of images and overfit on frontal images. 
While in practical applications, an ideal method should be \emph{pose and expression robust}, which is able to generate high-quality results even if source images and reference images show different poses and expressions. 
Secondly, the existing methods cannot perform customizable makeup transfer since they encode makeup styles into low dimension vectors which lose the spatial information. An ideal makeup transfer method need be capable of realizing \emph{partial} and \emph{shade-controllable} makeup transfer. Partial transfer indicates transferring the makeup of specified facial regions separately, e.g., eye shadows or lipstick. Shade-controllable transfer means the shade of the transferred makeup can be controllable from light to heavy.

To solve these challenges, we propose a novel Pose and expression robust Spatial-aware GAN, which consists of a Makeup Distill Network (MDNet), an Attentive Makeup Morphing (AMM) module and a Makeup Apply Network (MANet).
Different from the previous approaches that simply input two images into the  network or recombine makeup latent code and identity latent code to perform transfer, we design PSGAN to transfer makeup through scaling and shifting the feature map for only once, inspired by style transfer methods \cite{Huang2017ArbitraryST,Dumoulin2016ALR}.
Comparing with general style transfer, makeup transfer is more difficult since the human perception system is very sensitive to the artifacts on faces. Also, makeup styles contain subtle details in each facial region instead of general styles. To this end, we propose MDNet to disentangle the makeup from the reference image into two makeup matrices, i.e., the coefficient matrix $\gamma$ and bias matrix $\beta$ which both have the same spatial dimensions with visual features. These matrices embed the makeup information and serve as the shifting and scaling parameters. Then, $\gamma$ and $\beta$ are morphed and adapted to the source image by the AMM module which calculates an attentive matrix $A$ to produce adapted makeup matrices $\gamma^{\prime}$ and $\beta^{\prime}$. 
The AMM module utilizes the face parsing maps and facial landmarks to build the pixel-wise correspondences between source images and reference images, which solves the misalignment of faces. 
Finally, the proposed MANet conducts makeup transfer through applying pixel-wise multiplication and addition on visual features using $\gamma^{\prime}$ and $\beta^{\prime}$.


Since the makeup style has been distilled in a spatial-aware way, \emph{partial transfer} can be realized by applying masks pixel-wisely according to the face parsing results.
For example, in the top left panel of Figure \ref{p1}, the lip gloss, skin and eye shadow can be individually transferred from the reference image to the source image.
\emph{Shade-controllable transfer} can be realized through multiplying the weights of makeup matrices by coefficient within $[0,1]$.
As shown in the bottom left panel of Figure \ref{p1}, where the makeup shade is increasingly heavier. 
Moreover, the novel AMM module effectively assists the generating of  \emph{pose and expression robust} results, as shown in the right part of  Figure \ref{p1}. We also directly apply transfer to every frame of facial videos and still get nice and consistent results.
With the three novel components, PSGAN satisfies the requirements we pose for an ideal customizable makeup transfer method. 

We make the following contributions in this paper:
\begin{itemize}
   \item To our best knowledge, PSGAN is the first to simultaneously realize partial, shade-controllable, and pose/expression robust makeup transfer, which facilitates the applications in the real-world environment.
   \item A MDNet is introduced to disentangle the makeup from the reference image as two makeup matrices. The spatial-aware makeup matrices enable the flexible partial and shade-controllable transfer.
   \item An AMM module that adaptively morphs the makeup matrices to source images is proposed, which enables pose and expression robust transfer.
   \item A new Makeup-Wild dataset containing images with diverse poses and expressions is collected for better evaluations.
\end{itemize}

\begin{figure*}[!t]
    \centering
    \includegraphics[width=1.0\linewidth]{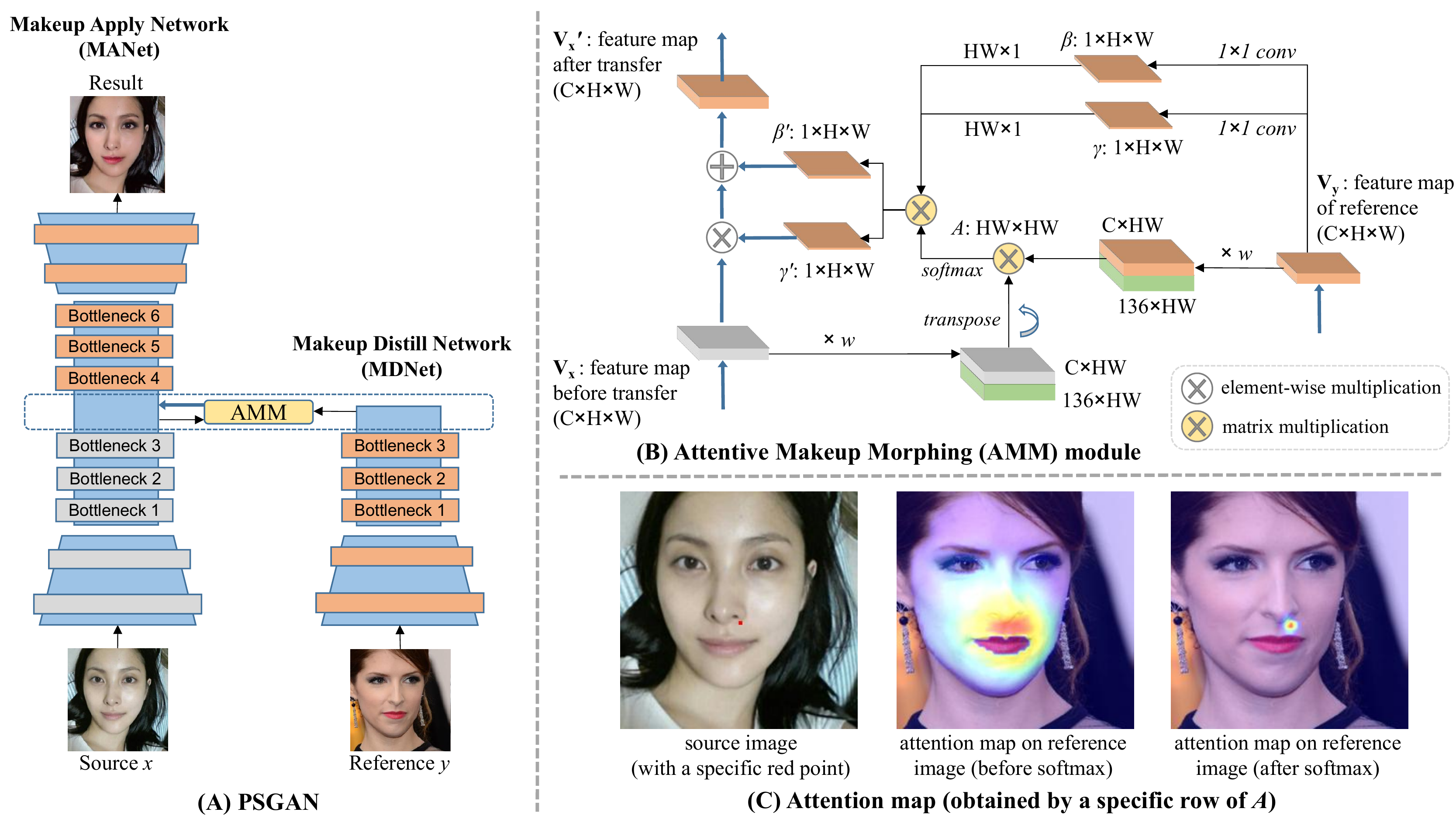}
    \caption{
        \textbf{(A)} Illustration of PSGAN framework. MDNet distills makeup matrices from the reference image. AMM module applies the adapted makeup matrices to the output feature map of the third bottleneck of MANet to achieve makeup transfer. 
        \textbf{(B)} Illustration of AMM module. Green blocks with 136 (68$\times$2) channels indicate relative position features of the pixels, which are then concatenated with $C$-channel visual features. Thus, the attention map is computed for each pixel in the source image through the similarity of relative positions and visual appearances. The adapted makeup matrices $\gamma^{\prime}$ and $\beta^{\prime}$ are produced by the AMM module, which are then multiplied and added to feature maps of MANet element-wisely. The orange and the gray blocks in the figure indicate visual features with makeup and without makeup. \textbf{(C)} Attention maps for a specific red point in the source image. Note that we only calculate attentive values for pixels that belong to the same facial region. Thus, there are no response values on the lip and eye of the reference image.
    }
    \vspace{-3mm}
    \label{framework}
\end{figure*} 

\section{Related Work}
\subsection{Makeup Transfer}
Makeup transfer has been studied a lot these years \cite{Tong2007ExampleBasedCT,Guo2009DigitalFM,Li2015SimulatingMT,Liu2016MakeupLA,Liu2014WowYA,Alashkar2017ExamplesRulesGD}.
BeautyGAN \cite{Li2018BeautyGANIF} first proposed a GAN framework with dual input and output for makeup transfer and removal simultaneously. They also introduced a makeup loss that matches the color histogram in different parts of faces for instance-level makeup transfer. BeautyGlow \cite{ChenBeautyGlow2019} proposed a similar idea on the Glow framework and decomposed makeup component and non-makeup component. PairedCycleGAN \cite{Chang2018PairedCycleGANAS} employed an additional discriminator to guide makeup transfer using pseudo transferred images generated by warping the reference face to the source face. LADN \cite{Gu2019LADNLA} leveraged additional multiple overlapping local discriminators for dramatic makeup transfer.
However, the above approaches often fail on transferring in-the-wild images and cannot adjust transfer precisely and partially, which limits their applications, such as the makeup transfer in videos.

\subsection{Style Transfer}

Style transfer has been investigated extensively \cite{Gatys2015ANA,Gatys2016PreservingCI,Johnson2016PerceptualLF,Luan2017DeepPS,Taigman2016UnsupervisedCI}. \cite{Gatys2016ImageST} proposed to derive image representations from CNN, which can be separated and recombined to synthesize images. Some methods are developed to solve the fast style transfer problem. \cite{Dumoulin2016ALR} found the vital role of normalization in style transfer networks and achieved fast style transfer by the conditional instance normalization. While their methods can only transfer a fixed set of styles and cannot adapt to arbitrary new styles. Then, \cite{Huang2017ArbitraryST} proposed adaptive instance normalization (AdaIN) that aligns the mean and variance of the content features with those of the style features and achieved arbitrary style transfer. Here, we propose spatial-aware makeup transfer for each pixel rather than transferring a general style from the reference.

\subsection{Attention Mechanism}
Attention mechanism has been utilized in many areas \cite{Xu2015ShowAA,Mnih2014RecurrentMO,Hu2017SqueezeandExcitationN,Rush2015ANA}. \cite{Vaswani2017AttentionIA} proposed the attention mechanism in the natural language processing area by leveraging a self-attention module to compute the response at a position in a sequence (e.g., a sentence) by attending to all positions and taking their weighted average in an embedding space.  \cite{Wang2017NonlocalNN} proposed the non-local network, which is to compute the response at a position as a weighted sum of the features at all positions. Inspired by these works, we explore the application of attention module by calculating the attention between two feature maps. Unlike the non-local network that only considers visual appearance similarities, our proposed AMM module computes the weighted sum of another feature map by considering both visual appearances and locations.
\vspace{-2mm}

\section{PSGAN}
\subsection{Formulation}

Let $X$ and $Y$ be the source image domain and the reference image domain. 
Also, we utilize $\left\{x^{n}\right\}_{n=1, \ldots, N}, x^{n} \in X$ and $\left\{y^{m}\right\}_{m=1, \ldots, M}, y^{m} \in Y$  to represent the examples of two domains respectively.  Note that paired datasets are not required. That is, the source and reference images have different identities. 
We assume $x$ is sampled from $X$ according to the distribution $\mathcal{P}_{X}$ and  $y$ is sampled from $Y$ according to the distribution $\mathcal{P}_{Y}$.
Our proposed PSGAN learns a transfer function $G: \{x, y\} \rightarrow \tilde{x}$, where the transferred image $\tilde{x}$ has the makeup style of the reference image $y$ and preserves the identity of the source image $x$.

\subsection{Framework}

\textbf{Overall.} The framework of PSGAN is shown in Figure \ref{framework} (A). 
Mathematically, it is formulated as $\tilde{x}= G(x, y)$.
It can be divided into three parts.
1) \emph{Makeup distill network.} MDNet extracts the makeup style from the reference image $y$ and represents it as two makeup matrices $\gamma$ and $\beta$, which have the same height and width as the feature map.
2)  \emph{Attentive makeup morphing module.}  Since source images and reference images may have large discrepancies in expressions and poses, the extracted makeup matrices cannot be directly applied to the source image $x$. We then propose an AMM module to morph the two makeup matrices to two new matrices $\gamma^{\prime}$ and $\beta^{\prime}$ which are adaptive to the source image by considering the similarities between pixels of the source and reference.
3) \emph{Makeup apply network.} The adaptive makeup matrices $\gamma^{\prime}$ and $\beta^{\prime}$ are applied to the bottleneck of the MANet to perform makeup transfer with pixel-level guidance by element-wise multiplication and addition.

\textbf{Makeup distill network.} The MDNet utilizes the encoder-bottleneck architecture used in  \cite{Choi2017StarGANUG} without the decoder part. 
It disentangles the makeup related features, e.g., lip gloss, eye shadows, from the intrinsic facial features, e.g., facial shape, the size of eyes.
The makeup related features are represented as two makeup matrices  $\gamma$ and $\beta$, which are used to transfer the makeup by pixel-level operations. As shown in Figure \ref{framework} (B), the output feature map of MDNet $\mathbf{V_y}\in \mathbb{R}^{C \times H \times W}$ is fed into two $1\times1$ convolution layers to produce $\gamma \in \mathbb{R}^{1\times H \times W}$ and $\beta \in \mathbb{R}^{1 \times H \times W}$, where $C$, $H$ and $W$ are the number of channels, height and width of the feature map.

\textbf{Attentive makeup morphing module.} 
Since the source and reference images may have different poses and expressions, the obtained spatial-aware $\gamma$ and $\beta$ cannot be applied directly to the source image. The proposed AMM module calculates an attentive matrix $A \in \mathbb{R}^{HW \times HW}$ to specify how a pixel in the source image $x$ is morphed from the pixels in the reference image $y$, where $A_{i,j}$ indicates the attentive value between the $i$-th pixel $x_i$ in image $x$ and the $j$-th pixel $y_j$ in image $y$.

Intuitively, makeup should be transferred between the pixels with similar relative positions on the face, and the attentive values between these pixels should be high. For example, the lip gloss region of the transferred result $\tilde{x}$ should be sampled from the corresponding lip gloss region of the reference image $y$.
To describe the relative positions, we take the facial landmarks as anchor points. The relative position feature of pixel $x_i$ is represented by $\mathbf{p}_i \in \mathbb{R}^{136}$, which is reflected in the differences of coordinates between pixel $x_i$  and $68$ facial landmarks, calculated by
\begin{equation}
    \begin{split}
    \mathbf{p}_i= [f(x_i)-f(l_1), f(x_i)-f(l_2), \dots, f(x_i) - f(l_{68}),\\ g(x_i)-g(l_1), g(x_i)-g(l_2), \dots, g(x_i) - g(l_{68})],
    \label{p}
    \end{split}
\end{equation}
where $f(\cdot)$ and $g(\cdot)$ indicate the coordinates on $x$ and $y$ axes, $l_i$ indicates the $i$-th facial landmark obtained by the 2D facial landmark detector \cite{Zhang2016JointFD}, which serves as the anchor point when calculating $\mathbf{p}_i$. In order to handle faces that occupy different sizes in images, we divide $\mathbf{p}$ by its two-norm (i.e., $\frac{\mathbf{p}}{\left\| \mathbf{p} \right\|}$) when calculating the attentive matrix.

Moreover, to avoid unreasonable sampling pixels with similar relative positions but different semantics, we also consider the visual similarities between pixels (e.g., $x_i$ and $y_j$), which are denoted as the similarities between $\mathbf{v}_i$  and $\mathbf{v}_j$ that extracted from the third bottleneck of MANet and MDNet respectively.
To make the relative position to be the primary concern, we multiply the visual features by a weight when calculating $A$.
Then, the relative position features are resized and concatenated with the visual features along the channel dimension.
As Figure \ref{framework} (B) shows, the attentive value $A_{i,j}$ is computed by considering the similarities of both visual appearances and relative positions via
\begin{equation}
A_{i, j} = \frac{\exp \left( [w\mathbf{v}_i, \frac{\mathbf{p}_i}{\left\| \mathbf{p}_i \right\|} ]^T [w\mathbf{v}_j, \frac{\mathbf{p}_j}{\left\| \mathbf{p}_j \right\|}]  \right) \mathbb{I}  (m^i_x = m^j_y) }{\sum_{j} \exp \left( [w\mathbf{v}_i, \frac{\mathbf{p}_i}{\left\| \mathbf{p}_i \right\|} ]^T [w\mathbf{v}_j, \frac{\mathbf{p}_j}{\left\| \mathbf{p}_j \right\|}] \right) \mathbb{I}  (m^i_x = m^j_y)},
\label{equ5}
\end{equation}
where $[\cdot, \cdot]$ denotes the concatenation operation, $\mathbf{v} \in \mathbb{R}^{C}$ and $\mathbf{p} \in \mathbb{R}^{136} $ indicate the visual features and relative position features, $w$ is the weight for visual features.
$\mathbb{I}(\cdot)$ is an indicator function whose value is $1$ if the inside formula is true, $m_x, m_y \in \{0, 1, \dots, N-1\}^{H \times W}$ are the face parsing map of source image $x$ and reference image $y$, where $N$ stands for the number of facial regions ($N$ is $3$ in our experiments including eyes, lip and skin), $m^i_x$ and $m^j_y$ indicate the facial regions that $x_i$ and $x_j$ belong to. Note that we only consider the pixels belonging to same facial region, i.e., $m^i_x = m^j_y$ , by applying indicator function $\mathbb{I}(\cdot)$.

Given a specific point that marked in red in the lower-left corner of the nose in the source image, 
the middle image of Figure \ref{framework} (C) shows its attention map by reshaping a specific row of the attentive matrix $A_{i,:} \in \mathbb{R}^{1 \times HW}$ to $H \times W$.
We can see that only the pixels around the left corner of the nose have large values. After applying softmax, attentive values become more gathered.
This verifies that our proposed AMM module is able to locate semantically similar pixels to attend.

We multiply attentive matrix $A$ by the $\gamma$ and $\beta$, and get the morphed makeup matrices $\gamma^{\prime}$ and $\beta^{\prime}$. 
More specifically, the matrices $\gamma^\prime$ and $\beta^\prime$ are computed by
\begin{equation}
\begin{split}
\gamma^\prime_i = \sum_{j} A_{i, j} \gamma_j ; \\
\beta^\prime_i = \sum_{j} A_{i, j} \beta_j ,
\label{equ4}
\end{split}
\end{equation}
where $i$ and $j$ are the pixel index of $x$ and $y$. After that, the matrix $\gamma^{\prime}  \in \mathbb{R}^{1 \times H \times W}$ and $\beta^{\prime} \in \mathbb{R}^{1 \times H \times W}$ are duplicated and expanded along the channel dimension to produce the makeup tensors $\Gamma^{\prime}  \in \mathbb{R}^{C \times H \times W}$ and $B^{\prime}  \in \mathbb{R}^{C \times H \times W}$, which will be the input of MANet.

\textbf{Makeup apply network.} MANet utilizes a similar encoder-bottleneck-decoder architecture as \cite{Choi2017StarGANUG}. As shown in Figure \ref{framework} (A), the encoder part of MANet shares the same architecture with MDNet, but they do not share parameters.
In the encoder part, we use instance normalizations that have no affine parameters that make the feature map to be a normal distribution.
In the bottleneck part, the morphed makeup tensors $\Gamma^{\prime}$ and $B^{\prime}$ obtained by the AMM module are applied to the source image feature map $\mathbf{V_x}\in \mathbb{R}^{C \times H \times W}$. 
The activation values of the transferred feature map $\mathbf{V_x}^{\prime}$ are calculated by
\begin{equation}
\mathbf{V_x}^{\prime} = \Gamma^{\prime} \mathbf{V_x} +B ^{\prime}.
\label{equ1}
\end{equation}
Eq. (\ref{equ1}) gives the function of makeup transfer. The updated feature map $\mathbf{V_x}^{\prime}$ is then fed to the subsequent decoder part of MANet to produce the transferred result.

\subsection{Objective Function}

\textbf{Adversarial loss.} We utilize two discriminators $D_X$ and $D_Y$ for the source image domain $X$ and the reference image domain $Y$, which try to discriminate between generated images and real images and thus help the generators synthesize realistic outputs. Therefore, the adversarial loss $L_D^{adv}$, $L_G^{adv}$ for discriminator and generator are computed by
\begin{equation}
    \begin{gathered}
        \begin{aligned} L_{D}^{adv} &= - \mathbb{E}_{x \sim \mathcal{P}_{X}}\left[\log D_{X}(x)\right] - \mathbb{E}_{y \sim \mathcal{P}_{Y}}\left[\log D_{Y}(y)\right] \\ &- \mathbb{E}_{x \sim \mathcal{P}_{X}, y \sim \mathcal{P}_{Y}}\left[\log \left(1-D_{X}(G(y, x))\right)\right] \\ &- \mathbb{E}_{x \sim \mathcal{P}_{X}, y \sim \mathcal{P}_{Y}}\left[\log \left(1-D_{Y}(G(x, y))\right)\right] 
        \end{aligned} \\
        \begin{aligned} L_{G}^{adv} = &- \mathbb{E}_{x \sim \mathcal{P}_{X}, y \sim \mathcal{P}_{Y}}\left[\log \left(D_{X}(G(y, x))\right)\right] \\ &- \mathbb{E}_{x \sim \mathcal{P}_{X}, y \sim \mathcal{P}_{Y}}\left[\log \left(D_{Y}(G(x, y))\right)\right] 
        \end{aligned}
    \end{gathered}
\end{equation}
 
\textbf{Cycle consistency loss.} Due to the lack of triplets data (source image, reference image, and transferred image), we train the network in an unsupervised way. Here, we introduce the cycle consistency loss proposed by \cite{Zhu2017UnpairedIT}. We use the L1 loss to constrain the reconstructed images and define the cycle consistency loss $L_G^{cyc}$ as
\begin{equation}
    \begin{aligned}
        L_G^{cyc} &=\mathbb{E}_{x \sim \mathcal{P}_{X}, y \sim \mathcal{P}_{Y}}\left[ \left\| G(G(x, y), x) - x \right\|_{1} \right]  \\ &+\mathbb{E}_{x \sim \mathcal{P}_{X}, y \sim \mathcal{P}_{Y}}\left[ \left\| G(G(y, x), y) - y \right\|_{1} \right].
    \end{aligned}
\end{equation}

\textbf{Perceptual loss.} When transferring the makeup style, the transferred image is required to preserve personal identity. Instead of directly measuring differences at pixel-level, we utilize a VGG-16 model pre-trained on ImageNet to compare the activations of source images and generated images in the hidden layer. Let $F_l(\cdot)$ denote the output of the $l$-th layer of VGG-16 model. We introduce the perceptual loss $L_G^{per}$ to measure their differences using L2 loss:
\begin{equation}
    \begin{aligned}
        L_G^{per} &=\mathbb{E}_{x \sim \mathcal{P}_{X}, y \sim \mathcal{P}_{Y}}\left[ \left\| F_l(G(x, y)) - F_l(x) \right\|_{2} \right]  \\ &+\mathbb{E}_{x \sim \mathcal{P}_{X}, y \sim \mathcal{P}_{Y}}\left[ \left\| F_l(G(y, x)) - F_l(y) \right\|_{2} \right].
    \end{aligned}
\end{equation}

\textbf{Makeup loss.} To provide coarse guidance for makeup transfer, we utilize the makeup loss proposed by \cite{Li2018BeautyGANIF}. Specifically, we perform histogram matching on the same facial regions of $x$ and $y$ separately and then recombine the results, denoted as $HM(x, y)$. As a kind of pseudo ground truth, $HM(x, y)$ preserves the identity of $x$ and has a similar color distribution with $y$. Then we calculate the makeup loss $L_G^{make}$ as coarse guidance by
\begin{equation}
    \begin{aligned}
        L_G^{make} &= \mathbb{E}_{x \sim \mathcal{P}_{X}, y \sim \mathcal{P}_{Y}}\left[ \left\| G(x, y) - HW(x, y) \right\|_{2} \right]  \\ &+\mathbb{E}_{x \sim \mathcal{P}_{X}, y \sim \mathcal{P}_{Y}}\left[ \left\| G(y, x) - HW(y, x) \right\|_{2} \right].
    \end{aligned}
\end{equation}

\textbf{Total loss.} The loss $L_D$ and $L_G$ for discriminator and generator of our approach can be expressed as
\begin{equation}
    \begin{gathered}
            L_D = \lambda_{adv} L_D^{adv} \\
            L_G = \lambda_{adv} L_G^{adv} + \lambda_{cyc} L_G^{cyc} + \lambda_{per} L_G^{per} + \lambda_{make} L_G^{make},
    \end{gathered} 
\end{equation}
where $\lambda_{adv}$, $\lambda_{cyc}$, $\lambda_{per}$, $\lambda_{make}$ are the weights to balance the multiple objectives. 

\section{Experiments}

\subsection{Data Collection}
Since the existing makeup datasets only consist of frontal facial images with neutral expressions, we collect a new Makeup-Wild dataset that contains facial images with various poses and expressions as well as complex backgrounds to test methods in the real-world environment. We collect data from the Internet and then manually remove images with frontal face or neutral expression. After that, we crop and resize the images to be $256 \times 256$ resolution without alignment.
Finally, $403$ with-makeup images and $369$ non-makeup images are collected to form the Makeup-Wild dataset.

\subsection{Experimental Setting and Details}
We train our network using the training part of the MT (Makeup Transfer) dataset \cite{Li2018BeautyGANIF,ChenBeautyGlow2019} and test it on the testing part of MT dataset and the Makeup-Wild dataset. MT dataset contains $1,115$ non-makeup images and $2,719$ with-makeup images which are mostly well-aligned, with the resolution of $361 \times 361$ and the corresponding face parsing results. We follow the splitting strategy of \cite{Li2018BeautyGANIF} to form the train/test set and conduct frontal face experiments in the test set of MT dataset since the examples in the test set are well-aligned frontal facial images. 
To further prove the effectiveness of PSGAN for handling pose and expression differences, we use the Makeup-Wild dataset as an extra test set. Note that we only train our network using the training part of the MT dataset for a fair comparison.

For all experiments, we resize the images to 256$\times$256, and utilize the $relu\_4\_1$ feature layer of VGG-16 for calculating perceptual loss. The weights of different loss functions are set as $\lambda_{adv} = 1$, $\lambda_{cyc} = 10$, $\lambda_{per} = 0.005$, $\lambda_{make} = 1$, and the weight for visual feature in AMM is set to $0.01$. We train the model for 50 epochs optimized by Adam \cite{kingma2014adam} with learning rate of 0.0002 and batch size of 1.

\subsection{Ablation Studies}

\begin{figure}[!t]
    \includegraphics[width=1\linewidth]{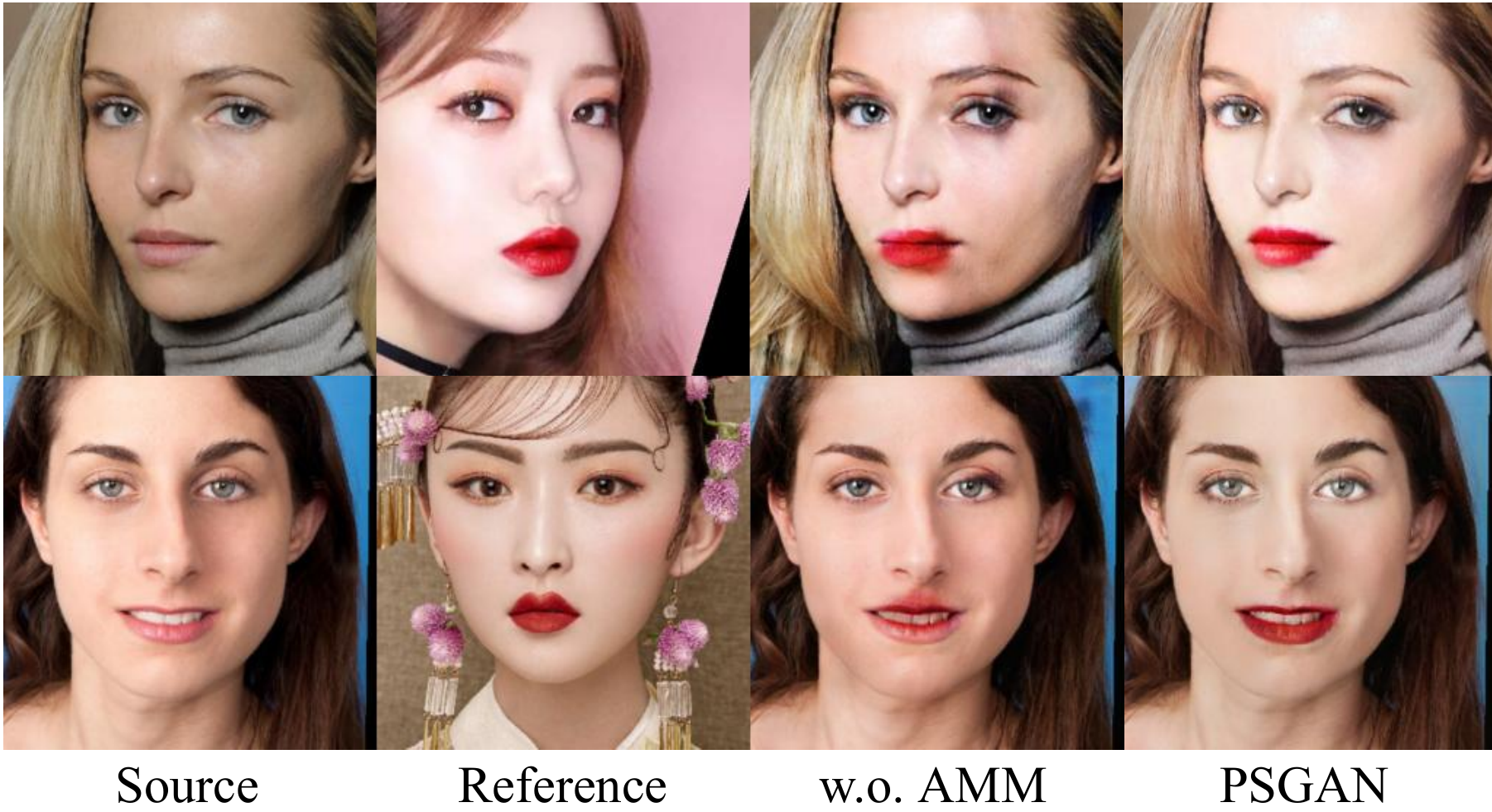}
    \caption{Without AMM module, the  makeup transfer results (the 3rd column) are bad due to pose and expression differences between source and reference images.}
    \label{ab1}
 \end{figure}

\textbf{Attentive makeup morphing module.}
In PSGAN,  AMM module  morphs the distilled makeup matrices $\gamma$ and $\beta$ to $\gamma^{\prime}$, $\beta^{\prime}$. It alleviates the pose and expression differences between source and reference images. 
The effectiveness of the AMM module is shown in Figure \ref{ab1}.
In the first row,  the pose of source and reference images are very different.  The bangs of the reference image are transferred to the skin of the source image without AMM. By applying AMM, the pose misalignment is well solved. 
A similar observation can be found in the second row: the expressions of source and reference images are smiling and neutral respectively, while the lip gloss is applied to the teeth region without the AMM module shown in the third column. After integrating AMM, lip gloss is applied to the lip region, bypassing the teeth area.
The experiments demonstrate that the AMM module can specify how a pixel in the source image is morphed from pixels of the reference instead of mapping the makeup from the same location directly. 

\begin{figure}[!t]
    \includegraphics[width=1\linewidth]{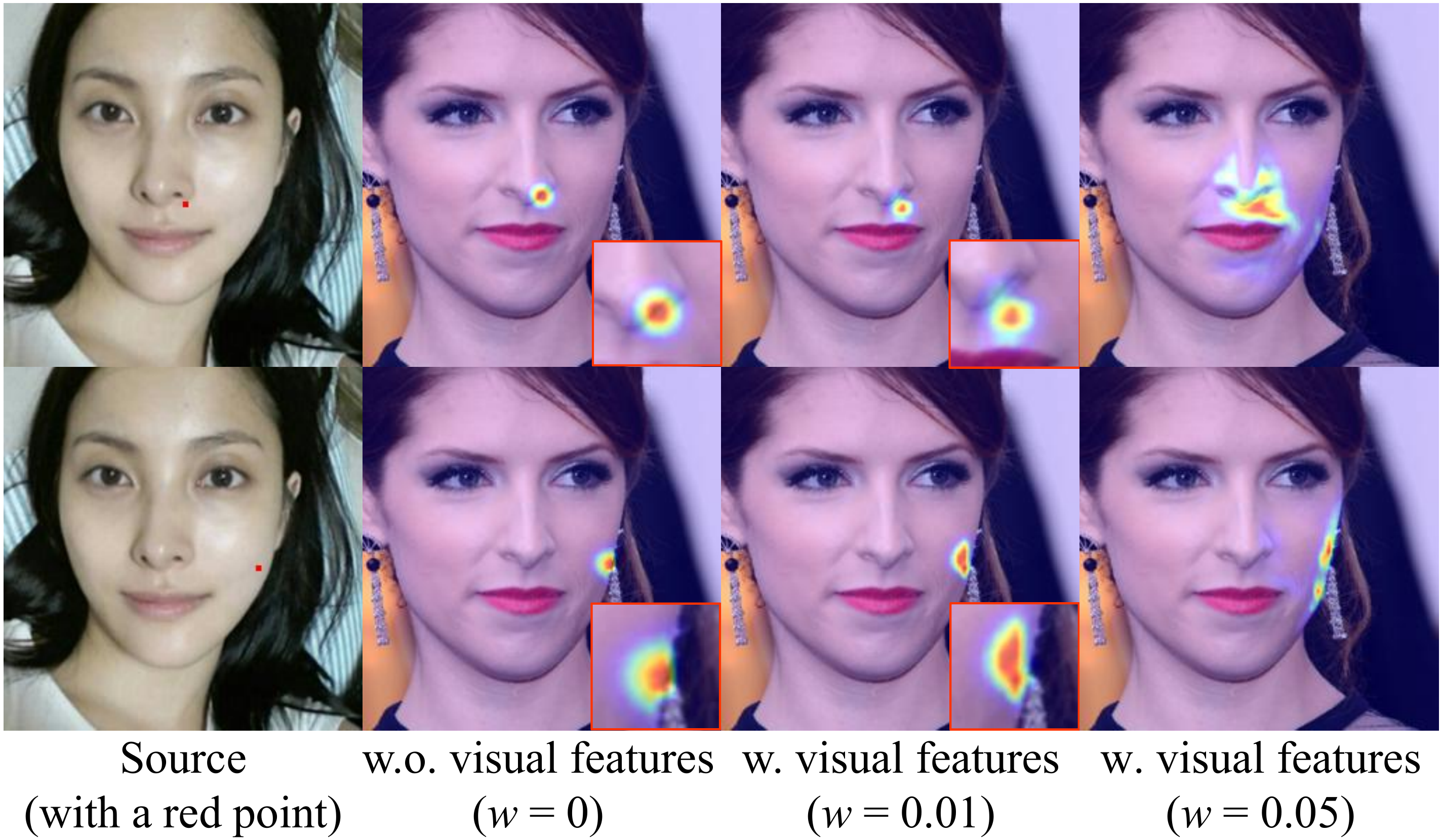}
    \caption{Given a red point on the skin, the corresponding attention maps with different weights on visual features are shown. Without using visual features, attention maps fail to avoid nostrils (1st row, 2nd column) and wrongly crosses the facial boundary (2nd row, 2nd column). While a larger weight leads to scattered and unreasonable attention maps.}
    \label{ab2}
 \end{figure}

\textbf{The weight of visual feature in calculating $A$.} In the AMM module, we calculate the attentive matrix $A$ by considering both the visual features $\mathbf{v}$ and relative positions $\mathbf{p}$ using Eq. (\ref{equ5}). 
Figure \ref{ab2} demonstrates that if only relative positions are considered by setting the weight to zero, the attentive maps in the second column are similar to a 2D Gaussian distribution.
In the first column of Figure \ref{ab2}, the red point on the skin of the source may wrongly receive makeup from the nostrils area in the reference image (1st row). The attention map also crosses the face boundary and covers the earrings (2nd row) which is unreasonable.
Besides, larger weights will lead to scattered and unreasonable attention maps, as shown in the last column.
After considering the appearance feature appropriately by setting the weight to $0.01$, the attention maps focus more on the skin and also bypass the nostrils as well as background.

 \begin{figure}[!t]
    \includegraphics[width=1\linewidth]{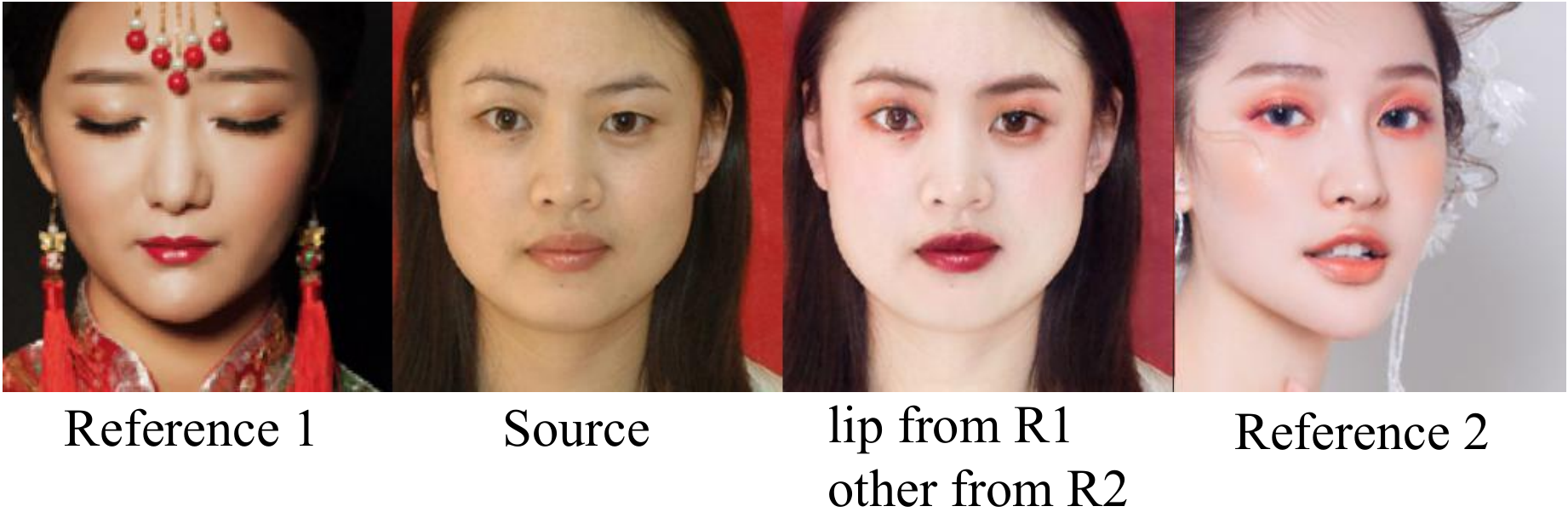}
    \caption{Given the source image (2nd column), the transferred images (3rd column) are generated by transferring the lipstick from reference 1 and other makeup from reference 2. }
    \label{partial}
 \end{figure}

\begin{figure*}
   \includegraphics[width=1\linewidth]{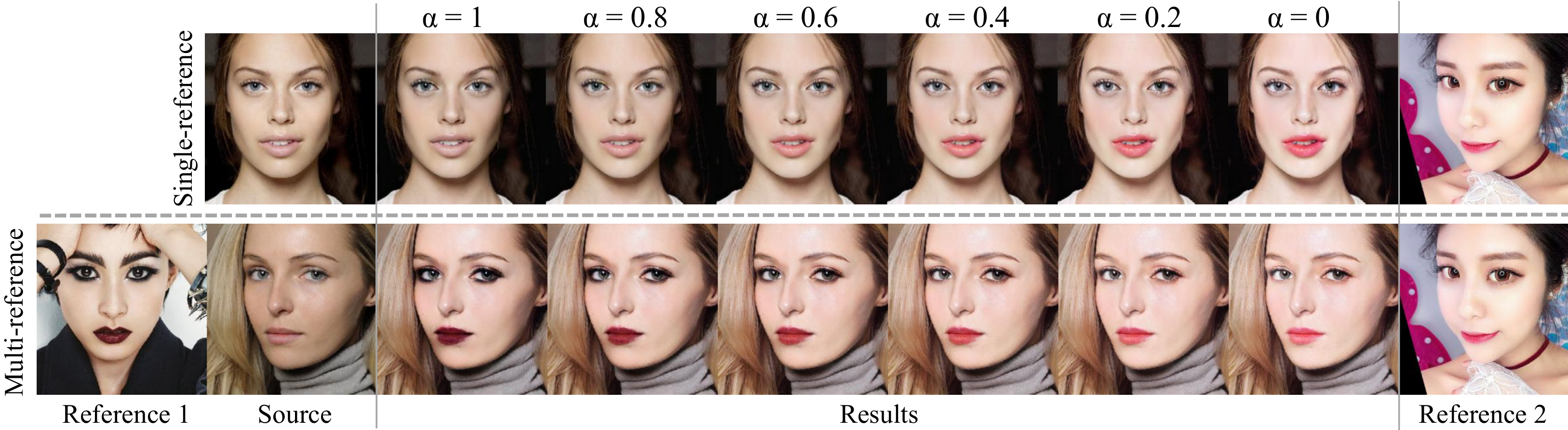}
   \caption{Results of interpolated makeup styles. If only one reference is used, adjusting the shade of makeup can be realized (1st row). If two references are used (1st column and last column), the makeup of the transferred images is gradually changing from reference 1 towards reference 2 from left to right (2nd rows). }
   \label{interpolated}
   \vspace{-2mm}
\end{figure*}

\subsection{Partial and Interpolated Makeup Transfer}
Since the makeup matrices $\gamma$ and $\beta$ are spatial-aware, the partial and interpolated transfer can be realized during testing.
To achieve partial makeup generation, we compute the new makeup matrices by weighting the matrices using the face parsing results.
Let $x$, $y_1$, and $y_2$ denote a source image and two reference images. We can obtain $\Gamma^{\prime}_x$, $B^{\prime}_x$ and $\Gamma^{\prime}_{y_1}$, $B^{\prime}_{y_1}$ as well as $\Gamma^{\prime}_{y_2}$, $B^{\prime}_{y_2}$ by feeding the images to MDNet.
In addition, we can obtain the face parsing mask $m_x$ of $x$ through the existing deep learning method \cite{Zhao2016PyramidSP}.
Suppose we want to mix the lipstick from $y_1$ and other makeup from $y_2$, we can first obtain the binary mask of the lip, denoted as $m^{l}_x \in \{0, 1\}^{H \times W}$.
Then, PSGAN can realize partial makeup transfer by assigning different makeup parameters on different pixels.
By modifying Eq. (\ref{equ1}), the partial transferred feature map $\mathbf{V_x}^{\prime}$ can be calculated by
    \begin{equation}
        \mathbf{V_x}^{\prime} = (m^{l}_x \Gamma^{\prime}_{y_1} + (1-m^{l}_x)\Gamma^{\prime}_{y_2}) \mathbf{V_x}  + (m^{l}_x B^{\prime}_{y_1}  + (1-m^{l}_x) B^{\prime}_{y_2}).
        \label{equ2}
        \end{equation}
Figure \ref{partial} shows the results by mixing the makeup styles from two references partially. The results on the third column recombine the makeup of lip from reference 1 and other part of makeup from reference 2, which are natural and realistic. Also, only transferring the lipstick from reference 1 and remain other parts unchanged can be achieved by assigning $x = y_2$. The new feature of partial makeup makes PSGAN realize the flexible partial makeup transfer.

Moreover, we can interpolate the makeup with two reference images by a coefficient $\alpha \in [0,1]$. We first get the makeup tensors of two references $y_1$ and $y_2$, and then compute the new parameters by weighting them  with the coefficient $\alpha$. The resulted feature map $\mathbf{V_x}^{\prime}$ is calculated by
\begin{equation}
\mathbf{V_x}^{\prime} = (\alpha \Gamma^{\prime}_{y_1} + (1-\alpha)\Gamma^{\prime}_{y_2}) \mathbf{V_x}  + (\alpha B^{\prime}_{y_1} + (1-\alpha) B^{\prime}_{y_2}).
\label{equ3}
\end{equation}
Figure \ref{interpolated} shows the interpolated makeup transfer results with one and two reference images. By feeding the new makeup tensors into MANet, we yield a smooth transition between two reference makeup styles. Similarly, we can adjust the shade of transfer using only one reference image by assigning $x = y_1$.
The generated results demonstrate that our PSGAN can not only control the shade of makeup transfer but also generate a new style of makeup by mixing the makeup tensors of two makeup styles.

We can also perform partial and interpolated transfer simultaneously by leveraging both the face parsing maps and coefficient thanks to the design of spatial-aware makeup matrices. The above experiments have demonstrated that PSGAN broadens the application range of makeup transfer significantly.

\begin{figure*}[!ht]
   \includegraphics[width=1\linewidth]{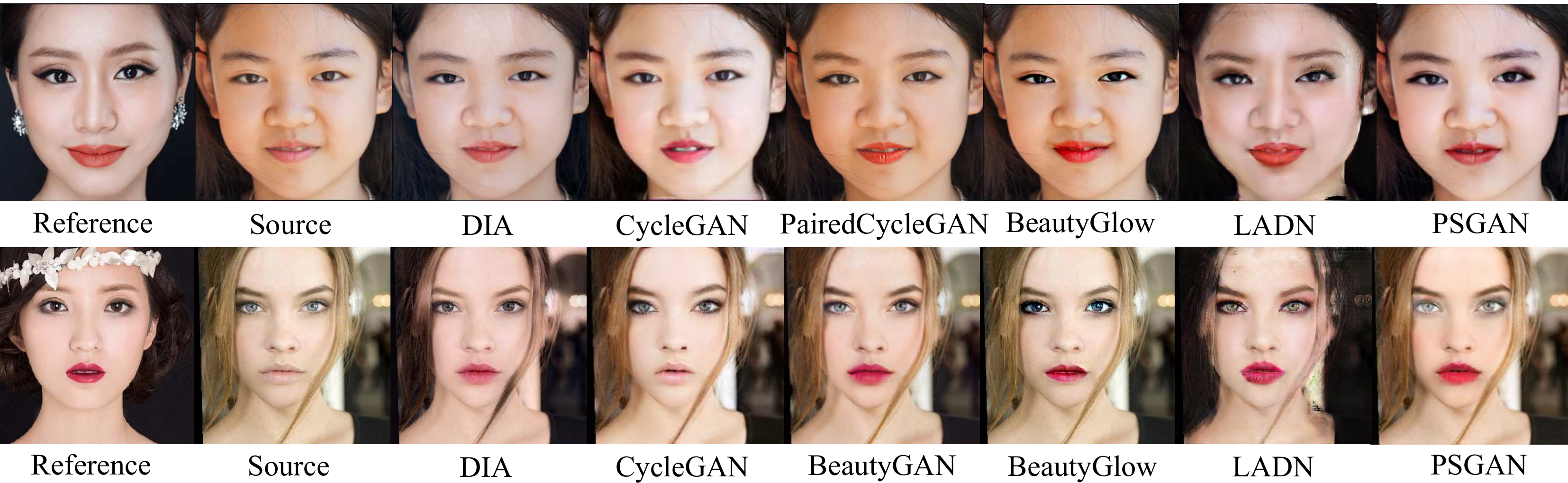}
   \caption{Qualitative comparison. PSGAN is able to generate realistic images with the same makeup styles as the reference.}
   \label{res1}
       \vspace{-2mm}
\end{figure*} 

\subsection{Comparison}
 
We conduct comparison with general image-to-image translation methos DIA \cite{Liao2017VisualAT} and CycleGAN \cite{Zhu2017UnpairedIT} as well as state-of-the-art makeup transfer methods BeautyGAN (BGAN) \cite{Li2018BeautyGANIF}, PairedCycleGAN (PGAN) \cite{Chang2018PairedCycleGANAS}, BeautyGlow (BGlow) \cite{ChenBeautyGlow2019} and LADN \cite{Gu2019LADNLA}. Current makeup transfer methods leverage face parsing maps \cite{Chang2018PairedCycleGANAS,ChenBeautyGlow2019,Li2018BeautyGANIF} and facial landmarks \cite{Gu2019LADNLA} for training and realize different functions as shown in Table \ref{t1}.
\begin{table}[!t]
    \centering
    \small
     
    \begin{tabular}{@{}c|ccc@{}}
        \toprule
        \multirow{2}{*}{Method} & \multicolumn{3}{c}{Functions} \\ \cmidrule(l){2-4} 
        & Shade    & Part    & Robust   \\ \midrule
        BGAN \cite{Li2018BeautyGANIF}    &       &      &    \\ \midrule
        PGAN \cite{Chang2018PairedCycleGANAS}     &       &      &    \\ \midrule
        BGlow \cite{ChenBeautyGlow2019}   & \checkmark &      &    \\ \midrule
        LADN \cite{Gu2019LADNLA}   & \checkmark &      &    \\ \midrule
        PSGAN   & \checkmark & \checkmark & \checkmark \\ \bottomrule
    \end{tabular}
    \vspace{1mm}
    \caption{Analysises of existing methods. ``Shade'', ``Part'' and ``Robust'' indicate shade-controllable, partial and pose/expression robust transfer respectively.}
    \label{t1}
\end{table}

\begin{table}[!t]
    \centering
    \begin{tabular}{@{}cccccc@{}}
      \toprule
      Test set & PSGAN & BGAN & DIA & CGAN & LADN\\ \midrule
      MT       &  \textbf{61.5}  &  32.5 &  3.25 & 2.5  &  0.25 \\ \midrule
      M-Wild   & \textbf{83.5} & 13.5 & 1.75 & 1.25 & 0.0 \\ \bottomrule
    \end{tabular}
    \vspace{1mm}
    \caption{Ratio selected as best (\%).}
    \label{t2}
    \vspace{-4mm}
\end{table}

\textbf{Quantitative Comparison.} We conduct a user study for quantitative evaluation on Amazon Mechanical Turk (AMT) that use BGAN, CGAN, DIA, and LADN as baselines. For a fair comparison, we only compare with methods whose code and pre-train model are released since we cannot guarantee a perfect re-implementation. We randomly select 20 source images and 20 reference images from both the MT test set and Makeup-Wild (M-Wild) dataset. After using the above methods to perform makeup transfer between these images, we obtain 800 images for each method. Then, 5 different workers are asked to choose the best images generated by five methods through considering image realism and the similarity with reference makeup styles. The generated images are shown in random order for a fair comparison. Table \ref{t2} shows the human evaluation results. Our PSGAN outperforms other methods by a large margin, especially on the M-Wild test set.

\textbf{Qualitative Comparison.} Figure \ref{res1} shows the qualitative comparison of PSGAN with other state-of-the-art methods on frontal faces in neutral expressions. Since the code of BeautyGlow and PairedCycleGAN is not released, we follow the strategy of BeautyGlow which cropped the results from corresponding papers. 
The result produced by DIA has an unnatural color on hair and background since it performs transfer in the whole image.
Comparatively, the result of CycleGAN is more realistic than that of DIA, but CycleGAN can only synthesize general makeup which is not similar to the reference.
The current makeup transfer methods outperform the previous methods. However, BeautyGlow fails to preserve the color of pupils and does not have the same foundation makeup as reference. We also use the pre-trained model released by the author of LADN, which produces blurry transfer results and unnatural background.
Compared to the baselines, our method is able to generate vivid images with the same makeup styles as reference.

\begin{figure}[!t]
    \includegraphics[width=1\linewidth]{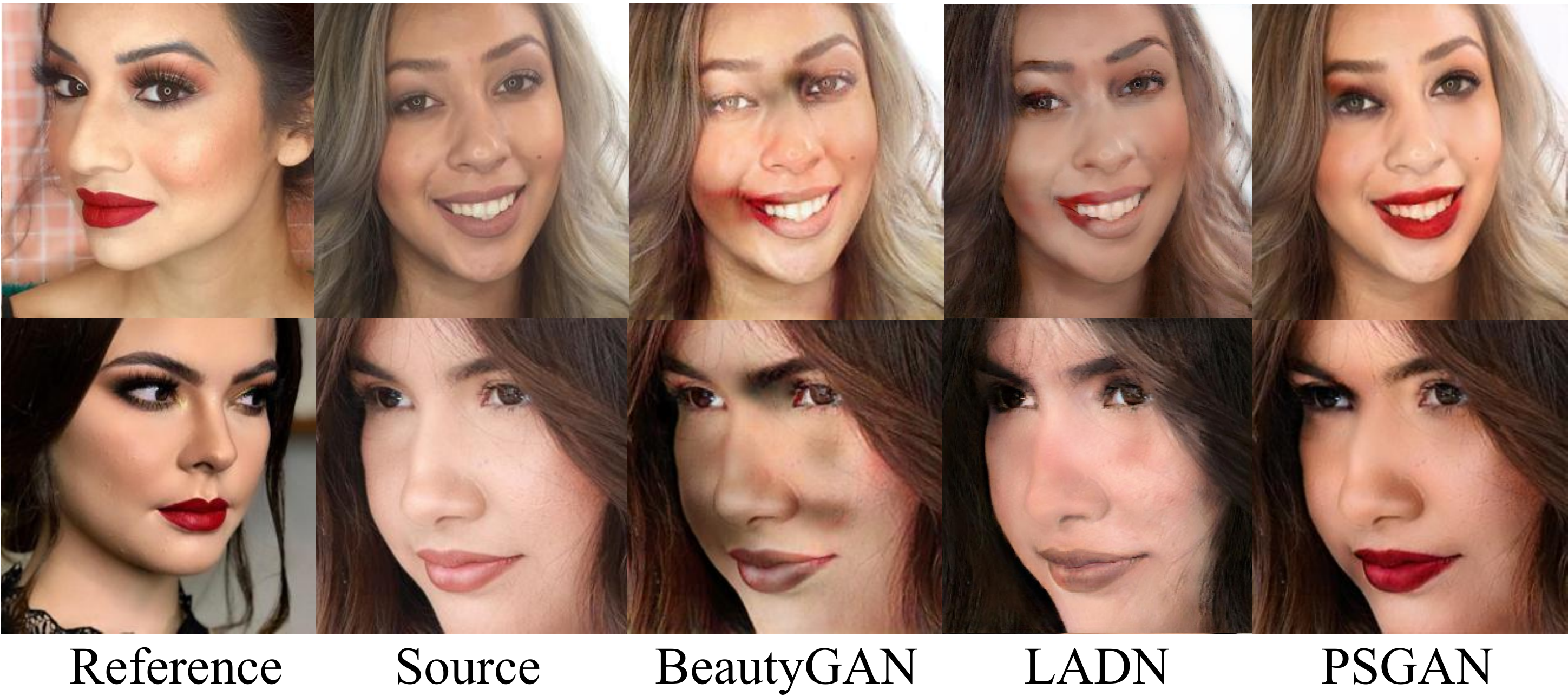}
    \caption{Qualitative comparison on M-Wild test set.}
    \label{wild}
    \vspace{-3mm}
\end{figure}

We also conduct a comparison on the M-Wild test set with the state-of-the-art method (BeautyGAN and LADN) that provide code and pre-trained model, as shown in Figure \ref{wild}.
Since the current methods lack an explicit mechanism to guide the direction of make transfer at the pixel-level and also overfit on frontal images, the makeup is applied in the wrong region of the face when dealing with images with different poses and expressions. For example, the lip gloss is transferred to the skin on the first row of Figure \ref{wild}. In the second row, other methods fail to perform transfer on faces with different sizes. However, the proposed AMM module can accurately assign the makeup for every pixel through calculating the similarities, which makes our results look better.

\subsection{Video Makeup Transfer}
\begin{figure}[!t]
   \includegraphics[width=1\linewidth]{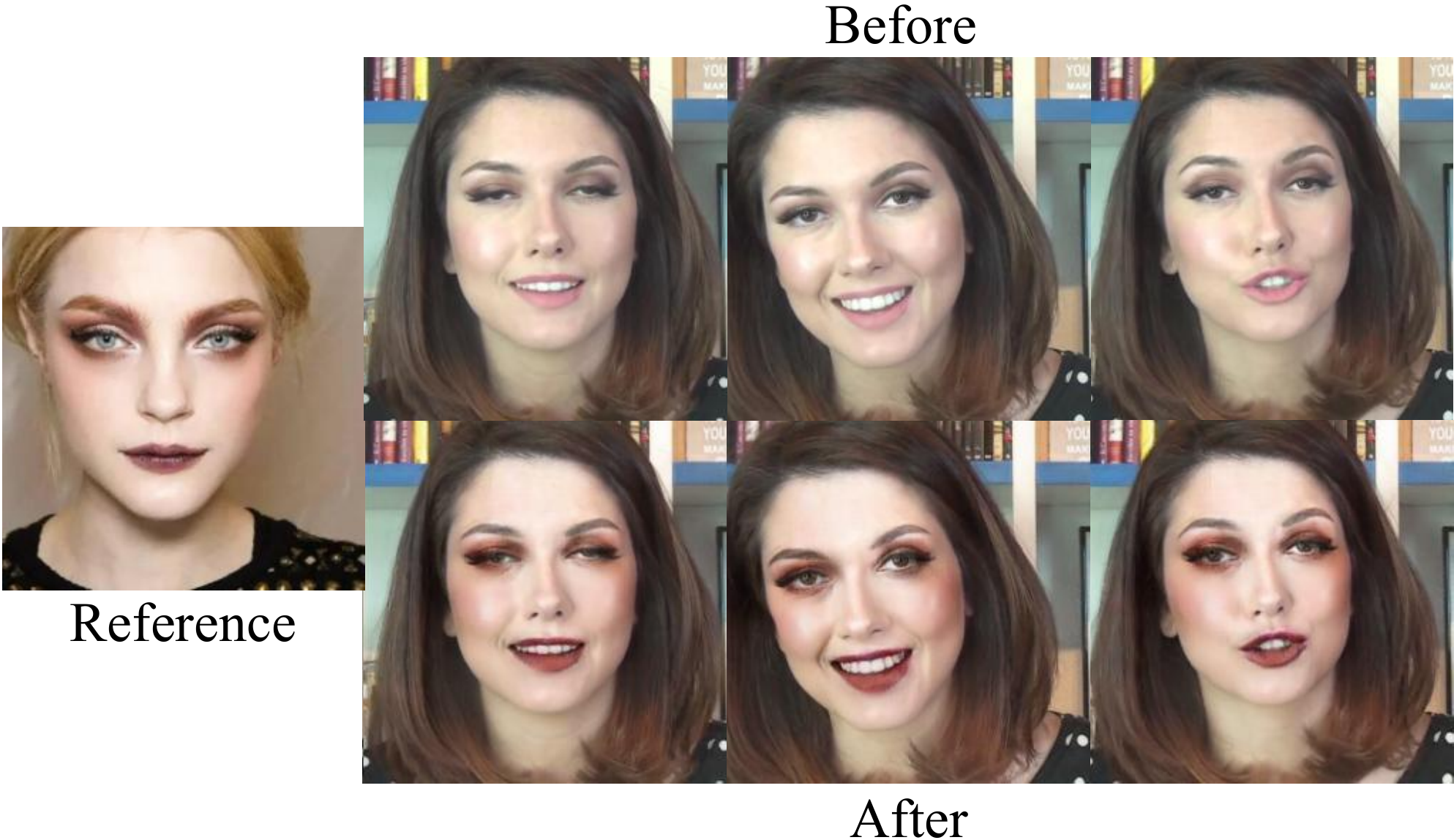}
   \caption{Video makeup transfer results of PSGAN.}
   \label{video}
   \vspace{-2mm}
\end{figure}

To transfer makeup for a person in the video is a challenging and meaningful task, which has wide prospects in the applications. However, the pose and expression of a face in the video are continuously changing which brings extra difficulties. To examine the effectiveness of our method, we simply perform makeup transfer on every frame of the video, as shown in Figure \ref{video}. By incorporating the design of PSGAN, we receive nice and stable transferred results.

\section{Conclusion}
In order to bring makeup transfer to real-world applications, we propose the Pose and expression robust Spatial-Aware GAN (PSGAN) that first distills the makeup style into two makeup matrices from the reference and then leverages an Attentive Makeup Morphing (AMM) module to conduct makeup transfer accurately. 
The experiments demonstrate our approach can achieve state-of-the-art transfer results on both frontal facial images and facial images that have various poses and expressions. Also, with the spatial-aware makeup matrices, PSGAN can transfer the makeup partially and adjust the shade of transfer, which greatly broadens the application range of makeup transfer. Moreover, we believe our novel framework can be used in other conditional image synthesis problems that require customizable and precise synthesis.

{\small
\bibliographystyle{ieee_fullname}
\bibliography{egbib}

\begin{thebibliography}{10}\itemsep=-1pt

\bibitem{Alashkar2017ExamplesRulesGD}
Taleb Alashkar, Songyao Jiang, Shuyang Wang, and Yun Fu.
\newblock Examples-rules guided deep neural network for makeup recommendation.
\newblock In {\em AAAI}, 2017.

\bibitem{Chang2018PairedCycleGANAS}
Huiwen Chang, Jingwan Lu, Fisher Yu, and Adam Finkelstein.
\newblock Pairedcyclegan: Asymmetric style transfer for applying and removing
  makeup.
\newblock In {\em CVPR}, 2018.

\bibitem{ChenBeautyGlow2019}
Hung-Jen Chen, Ka-Ming Hui, Sishui Wang, Li-Wu Tsao, Hong-Han Shuai, Wen-Huang
  Cheng, and National~Chiao Tung.
\newblock Beautyglow : On-demand makeup transfer framework with reversible
  generative network.
\newblock In {\em CVPR}, 2019.

\bibitem{Choi2017StarGANUG}
Yunjey Choi, Min-Je Choi, Munyoung Kim, Jung-Woo Ha, Sunghun Kim, and Jaegul
  Choo.
\newblock Stargan: Unified generative adversarial networks for multi-domain
  image-to-image translation.
\newblock In {\em CVPR}, 2017.

\bibitem{Dumoulin2016ALR}
Vincent Dumoulin, Jonathon Shlens, and Manjunath Kudlur.
\newblock A learned representation for artistic style.
\newblock {\em ArXiv}, abs/1610.07629, 2016.

\bibitem{Gatys2016PreservingCI}
Leon~A. Gatys, Matthias Bethge, Aaron Hertzmann, and Eli Shechtman.
\newblock Preserving color in neural artistic style transfer.
\newblock {\em ArXiv}, abs/1606.05897, 2016.

\bibitem{Gatys2015ANA}
Leon~A. Gatys, Alexander~S. Ecker, and Matthias Bethge.
\newblock A neural algorithm of artistic style.
\newblock {\em ArXiv}, abs/1508.06576, 2015.

\bibitem{Gatys2016ImageST}
Leon~A. Gatys, Alexander~S. Ecker, and Matthias Bethge.
\newblock Image style transfer using convolutional neural networks.
\newblock In {\em CVPR}, 2016.

\bibitem{Goodfellow2014GenerativeAN}
Ian~J. Goodfellow, Jean Pouget-Abadie, Mehdi Mirza, Bing Xu, David
  Warde-Farley, Sherjil Ozair, Aaron~C. Courville, and Yoshua Bengio.
\newblock Generative adversarial nets.
\newblock In {\em NeurIPS}, 2014.

\bibitem{Gu2019LADNLA}
Qiao Gu, Guanzhi Wang, Mang~Tik Chiu, Yu-Wing Tai, and Chi-Keung Tang.
\newblock Ladn: Local adversarial disentangling network for facial makeup and
  de-makeup.
\newblock In {\em ICCV}, 2019.

\bibitem{Guo2009DigitalFM}
Dong Guo and Terence Sim.
\newblock Digital face makeup by example.
\newblock In {\em CVPR}, 2009.

\bibitem{Hu2017SqueezeandExcitationN}
Jie Hu, Li Shen, and Gang Sun.
\newblock Squeeze-and-excitation networks.
\newblock In {\em CVPR}, 2017.

\bibitem{Huang2017ArbitraryST}
Xun Huang and Serge~J. Belongie.
\newblock Arbitrary style transfer in real-time with adaptive instance
  normalization.
\newblock In {\em ICCV}, 2017.

\bibitem{Johnson2016PerceptualLF}
Justin Johnson, Alexandre Alahi, and Li Fei-Fei.
\newblock Perceptual losses for real-time style transfer and super-resolution.
\newblock In {\em ECCV}, 2016.

\bibitem{kingma2014adam}
Diederik~P Kingma and Jimmy Ba.
\newblock Adam: A method for stochastic optimization.
\newblock {\em ArXiv}, abs/1412.6980, 2014.

\bibitem{Li2015SimulatingMT}
Chen Li, Kun Zhou, and Stephen Lin.
\newblock Simulating makeup through physics-based manipulation of intrinsic
  image layers.
\newblock In {\em CVPR}, 2015.

\bibitem{Li2018BeautyGANIF}
Tingting Li, Ruihe Qian, Chao Dong, Si Liu, Qiong Yan, Wenwu Zhu, and Liang
  Lin.
\newblock Beautygan: Instance-level facial makeup transfer with deep generative
  adversarial network.
\newblock In {\em ACM MM}, 2018.

\bibitem{Liao2017VisualAT}
Jing Liao, Yuan Yao, Lu Yuan, Gang Hua, and Sing~Bing Kang.
\newblock Visual attribute transfer through deep image analogy.
\newblock {\em ACM TOG}, 2017.

\bibitem{Liu2014WowYA}
Luoqi Liu, Junliang Xing, Si Liu, Hui Xu, Xi Zhou, and Shuicheng Yan.
\newblock "wow! you are so beautiful today!".
\newblock In {\em ACM MM}, 2013.

\bibitem{Liu2016MakeupLA}
Si Liu, Xinyu Ou, Ruihe Qian, Wei Wang, and Xiaochun Cao.
\newblock Makeup like a superstar: Deep localized makeup transfer network.
\newblock In {\em IJCAI}, 2016.

\bibitem{Luan2017DeepPS}
Fujun Luan, Sylvain Paris, Eli Shechtman, and Kavita Bala.
\newblock Deep photo style transfer.
\newblock In {\em CVPR}, 2017.

\bibitem{Mnih2014RecurrentMO}
Volodymyr Mnih, Nicolas Manfred~Otto Heess, Alex Graves, and Koray Kavukcuoglu.
\newblock Recurrent models of visual attention.
\newblock In {\em NeurIPS}, 2014.

\bibitem{Rush2015ANA}
Alexander~M. Rush, Sumit Chopra, and Jason Weston.
\newblock A neural attention model for abstractive sentence summarization.
\newblock In {\em EMNLP}, 2015.

\bibitem{Taigman2016UnsupervisedCI}
Yaniv Taigman, Adam Polyak, and Lior Wolf.
\newblock Unsupervised cross-domain image generation.
\newblock {\em ICLR}, 2016.

\bibitem{Tong2007ExampleBasedCT}
Wai-Shun Tong, Chi-Keung Tang, Michael~S. Brown, and Ying-Qing Xu.
\newblock Example-based cosmetic transfer.
\newblock In {\em PG}, 2007.

\bibitem{Vaswani2017AttentionIA}
Ashish Vaswani, Noam Shazeer, Niki Parmar, Jakob Uszkoreit, Llion Jones,
  Aidan~N. Gomez, Lukasz Kaiser, and Illia Polosukhin.
\newblock Attention is all you need.
\newblock In {\em NeurIPS}, 2017.

\bibitem{Wang2017NonlocalNN}
Xiaolong Wang, Ross~B. Girshick, Abhinav Gupta, and Kaiming He.
\newblock Non-local neural networks.
\newblock In {\em CVPR}, 2017.

\bibitem{Xu2015ShowAA}
Kelvin Xu, Jimmy Ba, Ryan Kiros, Kyunghyun Cho, Aaron~C. Courville, Ruslan
  Salakhutdinov, Richard~S. Zemel, and Yoshua Bengio.
\newblock Show, attend and tell: Neural image caption generation with visual
  attention.
\newblock In {\em ICML}, 2015.

\bibitem{Zhang2016JointFD}
Kaipeng Zhang, Zhanpeng Zhang, Zhifeng Li, and Yu Qiao.
\newblock Joint face detection and alignment using multitask cascaded
  convolutional networks.
\newblock {\em Signal Processing Letters}, 2016.

\bibitem{Zhao2016PyramidSP}
Hengshuang Zhao, Jianping Shi, Xiaojuan Qi, Xiaogang Wang, and Jiaya Jia.
\newblock Pyramid scene parsing network.
\newblock In {\em CVPR}, 2016.

\bibitem{Zhu2017UnpairedIT}
Jun-Yan Zhu, Taesung Park, Phillip Isola, and Alexei~A. Efros.
\newblock Unpaired image-to-image translation using cycle-consistent
  adversarial networks.
\newblock In {\em ICCV}, 2017.

\end{thebibliography}
}

\end{document}